%% file: eccv2022submission.tex
\documentclass[runningheads]{llncs}
\usepackage{graphicx}

\usepackage{comment}
\usepackage{amsmath,amssymb} 
\usepackage{color}
\usepackage[accsupp]{axessibility}  

\usepackage[mathscr]{euscript}  
\usepackage{booktabs}
\usepackage{enumitem}
\usepackage{cuted}
\usepackage{flushend}
\usepackage{siunitx}
\usepackage{adjustbox}
\usepackage{array}
\usepackage{multirow}
\usepackage{multicol}
\usepackage{microtype}
\usepackage[resetlabels]{multibib}
\usepackage{cuted}
\usepackage[font={footnotesize}]{caption}
\usepackage{bm}
\usepackage{url}

\usepackage{subcaption}
\usepackage[acronym,smallcaps]{glossaries}
\include{acronyms}

\usepackage{makecell}

\usepackage{xcolor}
\definecolor{dark-green}{RGB}{12,80,12}

\usepackage[pagebackref,breaklinks,colorlinks]{hyperref}
\usepackage{siunitx}
\usepackage[ruled,vlined]{algorithm2e}
\usepackage{algorithmic}
\usepackage{microtype}

\usepackage{caption}
\usepackage[capitalize]{cleveref}
\crefname{section}{Sec.}{Secs.}
\Crefname{section}{Section}{Sections}
\Crefname{table}{Table}{Tables}
\crefname{table}{Tab.}{Tabs.}
\crefname{algorithm}{Algo.}{Algos.}

\newcommand{\figref}[1]{Fig.~\ref{#1}}
\newcommand{\tabref}[1]{Tab.~\ref{#1}}
\newcommand{\secref}[1]{Sec.~\ref{#1}}

\newcommand{\equref}[1]{Eqn.~(\ref{#1})}

\newlist{todolist}{itemize}{2}
\setlist[todolist]{label=$\square$}

\begin{document}
\pagestyle{headings}
\mainmatter
\def\ECCVSubNumber{6229}  

\title{On Hyperbolic Embeddings in Object Detection}

\titlerunning{On Hyperbolic Embeddings in 2D Object Detection}
%
\author{Christopher Lang\inst{1,2} \and
Alexander Braun\inst{1} \and
Abhinav Valada\inst{2}}
\authorrunning{C. Lang et al.}
%
\institute{Robert Bosch GmbH
\and
University of Freiburg\\
\email{\{lang,valada\}@cs.uni-freiburg.de}}
\maketitle

\begin{abstract}
Object detection, for the most part, has been formulated in the euclidean space, where euclidean or spherical geodesic distances measure the similarity of an image region to an object class prototype. 
In this work, we study whether a hyperbolic geometry better matches the underlying structure of the object classification space.
We incorporate a hyperbolic classifier in two-stage, keypoint-based, and transformer-based object detection architectures and evaluate them on large-scale, long-tailed, and zero-shot object detection benchmarks.
In our extensive experimental evaluations, we observe categorical class hierarchies emerging in the structure of the classification space, resulting in lower classification errors and boosting the overall object detection performance.
\end{abstract}

\section{Introduction}
\label{sec:intro}
\input{sections/introduction}

\section{Related Work}
\label{sec:relatedwork}
\input{sections/related_work}

\section{Technical Approach}
\label{sec:technical}
\input{sections/system_overview}

\section{Experimental Evaluation}
\label{sec:experiments}

\input{sections/experiments}

\section{Conclusions}
\label{sec:conclusions}
\input{sections/conclusion}


{\small
\bibliographystyle{splncs04}
\bibliography{references.bib}
}

\appendix
\include{sections/supplementary}

\end{document}

%% file: acronyms.tex
\newacronym{rpn}{RPN}{Region Proposal Network}
\newacronym{roi}{RoI}{Region of Interest}
\newacronym{iou}{IoU}{intersection over union}
\newacronym{nms}{NMS}{non-maximum suppression}
\newacronym{rgcn}{R-GCNs}{Relational Graph Convolutional Networks \cite{Schlichtkrull2018}}
\newacronym{fpn}{FPN}{Feature Proposal Network}
\newacronym{kern}{KERN}{Knowledge-Embedded Routing Network \cite{Chen2019}}
\newacronym{bert}{BERT}{Bidirectional Encoder Representations from Transformers \cite{Devlin2019}}
\newacronym{gru}{GRU}{Gated Recurrent Cells \cite{Li2016}}
\newacronym{fcn}{FCN}{fully-connected network}
\newacronym{rcgn}{RCGN}{Relational Graph Neural Network}
\newacronym{ggnn}{GGNN}{Gated-Graph Neural Network}
\newacronym{mlp}{MLP}{Multi-Layer Perceptron}
\newacronym{coco}{COCO 2017}{Common Objects in Context \cite{Lin2014}}
\newacronym{lvis}{LVIS v1}{LVIS v1~\cite{gupta2019lvis}}
\newacronym{svg}{SVG}{singular value decomposition}
\newacronym{kge}{KGE}{knowledge graph embedded}
\newacronym{sota}{SoTA}{state-of-the-art}
\newacronym{mlr}{MLR}{multi logistic regression}

%% file: sections/introduction.tex
The object detection task entails localizing and classifying objects in an image.
Deep learning methods originate from the two-stage approach~\cite{ren2015:faster-rcnn,Cai2018,gosala2021bird} that uses encoded visual features of an input image to first search for image regions that potentially contain objects, and in a second step classifies and refines each proposal region in isolation. Over the years, alternate approaches have challenged these initial design choices: end-to-end formulations~\cite{redmon2018yolov3,Lin2020,Zhou2019b,Zhu2020,Carion2020} yield a real-time and fully-differentiable alternative to the two-stage approach, multi-resolution detectors~\cite{ssd} boost the detection accuracy for small objects, and learnable proposals~\cite{Sun2020} supersede the need for dense candidate regions. Even novel object detection paradigms have emerged, such as anchorless~\cite{Zhou2019b} and set-based methods~\cite{Carion2020,Sun2020} that bypass non-maximum suppression, and attention-based methods~\cite{Carion2020} that explicitly learn context between object proposals. 

However, many design paradigms have never been questioned, first and foremost, the learnable class prototypes as well as the euclidean embedding space in the classifier head. 
Recently, alternative classification space formulations such as hyperbolic embeddings have outperformed their euclidean counterparts in an increasing number of domains that underlie hierarchical representations, for instance, in graph embeddings~\cite{nickel2017poincare}, document clustering~\cite{NEURIPS2019_043ab21f}, and even image classification~\cite{khrulkov2020hyperbolic}. These successes are attributed to the exponentially growing distance ratio of hyperbolic spaces, which enables them to match the rate of growth in tree-like structures, where the number of nodes increases exponentially with the depth of the tree. Khrulkov~\textit{et~al.}~\cite{khrulkov2020hyperbolic} observed that the image embeddings generated from state-of-the-art feature extractors on image classification datasets have low $\delta$-hyperbolicity values, which implies a tree-likeness of the learned metric space. They suggest that these embeddings exhibit hyperbolic properties and outperform euclidean embeddings on image classification and person re-identification tasks.

\begin{figure}
	\centering
		\centering
		\includegraphics[width=0.8\textwidth]{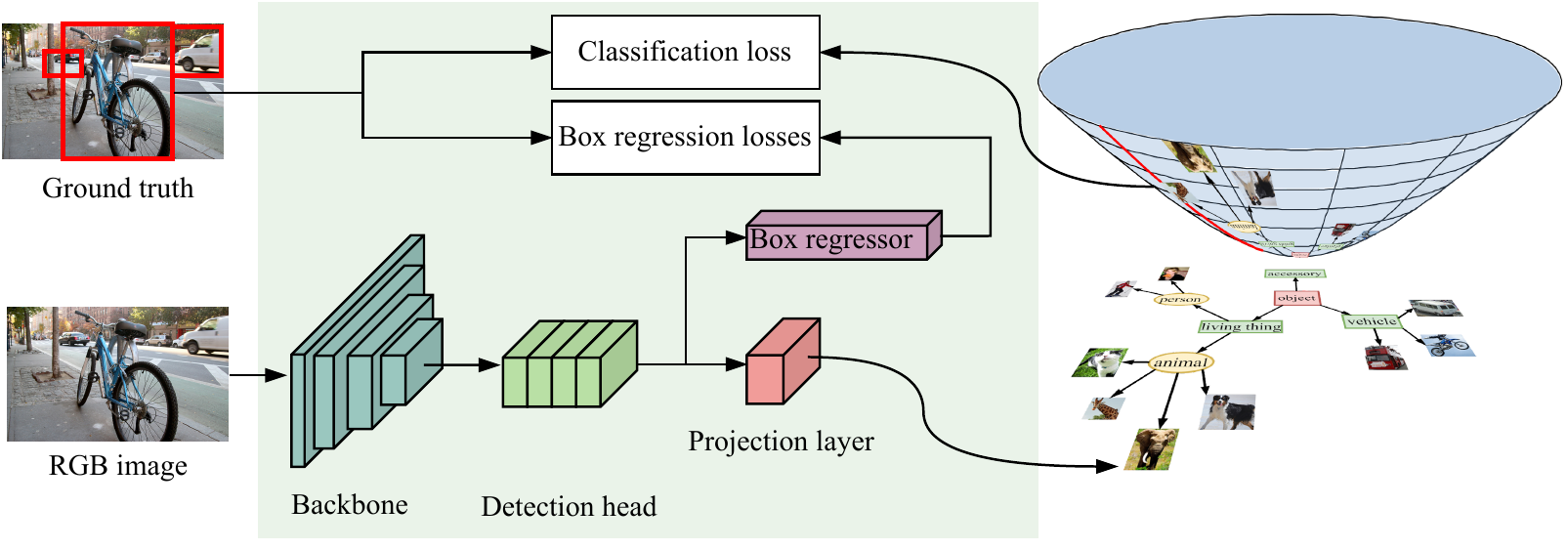}
	\caption{General object detection architecture using classification scores embedding in the hyperboloid model. It is built on object proposal features from an arbitrary detection neck, for instance a transformer-decoder or RoI head, using Euclidean operations. Finally, it outputs the classification logits computed in the learned hyperbolic metric space, i.e. calculates hyperbolic distances (\equref{eq:hyperbolic_distance}) to the learned class prototypes on the hyperboloid.}
	\vspace{-0.2cm}
\end{figure}

In this work, we enable the classification head of object detectors to learn hierarchical representations of visual features with low distortion in hyperbolic space.  We incorporate a hyperbolic classification head into various object detectors including Sparse R-CNN~\cite{Sun2020}, CenterNet~\cite{Zhou2019b}, and Deformable DETR~\cite{Zhu2020}, and evaluate the performance on closed-set object detection as well as long-tailed and zero-shot object detection. We analyze how it copes with the task to distinguish foreground and background representations and its interaction with localization. We observe a latent class hierarchy emerging from visual features, resulting in fewer and better classification errors, while simultaneously boosting the overall object detection performance. 

\noindent In summary, our main contributions in this work are:
\begin{itemize}[topsep=0pt,noitemsep]
    \item Formulation of a hyperbolic classification head for two-stage, keypoint-based, and transformer-based multi-object detection architectures.
    \item Evaluation of hyperbolic classification on closed-set, long-tailed, and zero-shot object detection.
    \item Analysis of the Hubness phenomenon in hyperbolic object detection.
\end{itemize}

%% file: sections/related_work.tex
In this section, we review the relevant works in the research area of object detection architectures, with special interest in their classification heads, as well as hyperbolic embeddings in the image domain.

{\parskip=5pt
\noindent\textit{Object Detection}: The first deep object detectors were anchor-based methods~\cite{ren2015:faster-rcnn,Cai2018,Sun2020,sirohi2021efficientlps} in which a \acrfull{rpn} first proposes image regions containing objects from a set of search anchors. In a second stage, a \acrfull{roi} head extracts visual features, from which it regresses enclosing bounding boxes and predicts a class label from a multi-set of classes for each object proposal in isolation. The model is trained by matching the groundtruth bounding boxes to object proposal boxes using a heuristic based on \acrfull{iou} matching alone. The inner product of learned class prototypes and the visual features yield classification scores that are optimized using a cross-entropy loss across all the classes, assuming a mutual exclusion of labels. Nowadays, a focal loss~\cite{Lin2020} formulation is more widely employed. It down-weights well-classified examples in the negative log-likelihood computation to mitigate the negative effect of class imbalance in the datasets. Due to its class-agnostic \acrfull{rpn}, such methods are nowadays widely applied in zero-shot object detection in which the objective is to map visual features to semantic embeddings, such that the methods can leverage their semantic context to recognize object types that were excluded from the training~\cite{zheng2021zero,Rahman2020,lang2021contrastive}.

Anchor-less detectors such as keypoint-based or center-based methods~\cite{Zhou2019b,Law2020} embody an end-to-end approach that directly predicts objects as class-heatmaps. For training, these methods use a logistic regression loss with multi-variate Gaussian distributions that are centered at each groundtruth box. Additional regression heads then deal with the estimation of bounding box offset and height/width \cite{Zhou2019b}, instance masks~\cite{lee2020centermask,mohan2022amodal} or temporal correspondences~\cite{hurtado2020mopt} in an image sequence. Zhang~\textit{et~al.}~\cite{zhang2020bridging} revealed that the performance gap between anchor-based and anchor-free detection can actually be reduced by unifying the definition of positive and negative training samples.

In recent years, an increasing number of works employ set-based training objective functions~\cite{Carion2020,Sun2020,Zhu2020,dai2021dynamic} that compute loss functions from a one-to-one matching between groundtruth and proposal boxes.  This objective is widely used with transformer-based detection heads~\cite{Carion2020,Zhu2020,dai2021dynamic} which process the feature map as a sequence of image patches, allowing all the positions in an image to interact using an attention operation. Such methods detect objects using cross-attention between learned object queries and visual embedding keys, as well as self-attention between object queries to capture their interrelations in the scene context. This explicitly modeled spatial awareness and the set-based loss enables these methods to make non-maximum suppression fully differentiable and reduce inductive biases.
The main modifications from the initial design of Carion~\textit{et~al.}~\cite{Carion2020} were in reducing the quadratic complexity in the number of image patches used for self-attention using sparser querying strategies~\cite{Zhu2020,sun2021sparse} that also support multiscale feature maps, thereby improving performance on small objects.}

{\parskip=5pt
\noindent\textit{Hyperbolic Embeddings}: Hyperbolic geometry defines spaces with negative constant Gaussian curvature, exchanging the parallel line postulate in Euclidean geometry by a form of limiting parallels. As a result, the distance-ratio of hyperbolic spaces increases exponentially~\cite{bridson2013metric} - a property that is incorporated by recent works to capture tree-like parent-child relations within the data~\cite{nickel2017poincare}. There are multiple definitions of hyperbolic planes which satisfy the axioms of a hyperbolic geometry. Among those, the Poincaré ball model is the most common choice for embedding tasks \cite{nickel2017poincare,khrulkov2020hyperbolic,liu2020hyperbolic}. The Poincaré model has two favorable properties: an intuitive visualization in 2D and the constraints on its embedding space  can be expressed by the euclidean norm. Nickel~\textit{et~al.}~\cite{nickel2017poincare} pioneered the learning of tree-structured graph embeddings in the Poincaré ball, and showed that the resulting hyperbolic embeddings can capture latent hierarchies and surpass Euclidean embeddings in terms of generalization ability and representation capacity. Poincaré embeddings were introduced in the image domain by Khrulkov~\textit{et~al.}~\cite{khrulkov2020hyperbolic} for image classification and person re-identification tasks. They observed that mini-batches of image embeddings generated from ResNet-18~\cite{he2016deep} feature extractors on Mini-ImageNet~\cite{deng2009imagenet} form a hyperbolic group, as the embeddings' Caley-Graph is $\delta$-hyperbolic. Motivated by this observation, they mapped the penultimate layer of a ResNet-18~\cite{he2016deep} backbone onto the Poincaré ball and performed a hyperbolic multi-logistic regression, which they showed to achieve superior results compared to euclidean methods on 5-shot classification on Mini-ImageNet~\cite{deng2009imagenet}. For zero-shot image classification, \cite{liu2020hyperbolic} transformed image embeddings using a ResNet-101~\cite{he2016deep} into hyperbolic vectors on the Poincaré ball model and perform classification based on distances to Poincaré embeddings of WordNet relations~\cite{nickel2017poincare} as class prototypes. In doing so, they outperform all euclidean baselines models in terms of hierarchical precision on ImageNet.
In a later work, Nickel~\textit{et~al.}~\cite{nickel2018learning} found that the hyperboloid model, also called Lorentz model, learns embeddings for large taxonomies more efficiently and more accurately than Poincaré embeddings. Furthermore, the distance computation on the hyperboloid model is numerically stable, as it avoids the stereographic projection.

Building on embeddings in the hyperboloid model~\cite{nickel2018learning}, we are going to perform object detection with hyperbolic detection head for two-stage, keypoint-based, and transformer-based detectors.
In addition to image classification, this requires to localize objects and distinguish between foreground and background image regions.
} 


%% file: sections/system_overview.tex
Closed-set object detection architectures share the classification head of a linear euclidean layer design~\cite{ren2015:faster-rcnn,besic2022dynamic,valverde2021there} that projects visual feature vectors $\mathbf{v}_i \in \mathbb{R}^{D}$ onto a parameter matrix $\mathbf{W} \in \mathbb{R}^{D \times C}$ composed of $C$ class prototypes:
\begin{equation}
    p(y = c | \mathbf{v}_i) = \textit{softmax}( \mathbf{W}^T \mathbf{v}_i)_c.
\end{equation}

These parameters are learned with a one-hot loss formulation that incentivizes the resulting class prototypes to be pairwise orthogonal. We propose to learn class prototypes in hyperbolic space to embed latent hierarchies with lower distortion. 
In the remainder of this section, we introduce hyperbolic embeddings in \secref{sec:hyperbolic}, the modifications to the classification losses in \secref{sec:loss_formulation}, and describe the incorporation into two-stage detectors, keypoint-based method, and transformer-based architectures in \secref{sec:object_detector_integration}.

\subsection{Hyperbolic Embeddings}
\label{sec:hyperbolic}

In this work, we analyze the $n$-dimensional hyperboloid model $\mathbb{H}^{n} $, also called Lorentz model, as a classification space for object detection. It is one of the several isometric models of the hyperbolic space, i.e., a Riemannian manifold with constant negative curvature $\kappa<0$. The non-zero curvature, however, is desirable to encode tree-like structures that capture latent hierarchical information due to the exponential growth of volume with distance to a point~\cite{bridson2013metric}. The limit case with $\kappa=0$ would recover behaviors as in Euclidean geometry, and hence the baseline object detectors.

The $n$-dimensional hyperboloid model $\mathbb{H}^{n}$ presents points on the upper sheet of a two-sheeted $n$-dimensional hyperboloid. It is defined by the Riemannian manifold $\mathcal{L} = (\mathbb{H}^n, \mathbf{g}_l)$ with
\begin{equation}
    \mathbb{H}^n = \{ \mathbf{x} \in \mathbb{R}^{n+1}: \langle \mathbf{x}, \mathbf{x} \rangle_{l} = -1, x_0 > 0  \},
\end{equation}
and 
\begin{equation}
    g_l(\mathbf{x}) = \text{diag}(-1, 1, \dots, 1) \in \mathbb{R}^{n+1 \times n+1},
\end{equation}
such that the Lorentzian scalar product $\langle \cdot, \cdot \rangle_{l}$ is given by
\begin{equation}
    \langle \mathbf{x}, \mathbf{y} \rangle_{l} = -x_0 y_0 + \sum_{i=1}^{n} x_n y_n, \quad \mathbf{x}, \mathbf{y} \in \mathbb{R}^{n+1}.
\end{equation}

To transform visual features $\mathbf{v} \in \mathbb{R}^{n+1}$ - extracted from euclidean modules like convolutional layers or \acrshort{mlp} - into points on the hyperboloid, we apply the exponential map as follows:
\begin{equation}
    \exp^k_0 (\mathbf{v}) = \sinh \left( \| \mathbf{v} \| \right) \frac{\mathbf{v}}{\| \mathbf{v} \|}.
\end{equation}
The distance between two points on the hyperboloid is then given by
\begin{equation}
    d_{\mathbb{H}^n}(\mathbf{x}_i, \mathbf{t}_c) = \text{arccosh}\left( -\langle \mathbf{x}_i, \mathbf{t}_c\rangle_{l} \right), \quad \mathbf{x}_i,  \mathbf{t}_c \in \mathbb{H}^{n}.
    \label{eq:hyperbolic_distance}
\end{equation}

When performing gradient optimization over the tangent space at a hyperbolic embedding $\mathbf{x} \in \mathbb{H}^{n}$, we employ the exponential map at $\mathbf{x}$ to the tangent vector, as shown in \cite{wilson2018gradient}.

\subsection{Focal Loss Integration}
\label{sec:loss_formulation}

The state-of-the-art object detectors compared in this work use a focal classification loss~\cite{Lin2020} that relies on the sigmoid function for computing the binary cross entropy.
Since distance metrics only span the non-negative real numbers, we compute logits by shifting the distances by a negative value as
\begin{equation}
    s_{i, c} = \Delta - \frac{\Delta}{d_{min}} d_{i, c},
    \label{eq:dist_to_scores}
\end{equation}
where $s_{i, c}$ is the classification score for proposal $i$ and class $c$, $d_{\mathbb{H}^n}(\mathbf{x}_i, \mathbf{t}_c)$ is the respective distance between the transformed visual feature vector and the prototype for class $c$, $\Delta$ is an offset that shifts the active regions of the activation function, and a scaling-hyperparameter $d_{min}$. We choose $\Delta \approx 1.4$ as the maximum value of the second derivative on the sigmoid, following the backpropagation tricks suggested by LeCun~\textit{et~al.}~\cite{lecun2012efficient}. The scaling parameter $d_{min}$ defines the distance that accounts for a classification confidence of $p=0.5$. It is set to the minimum inter-class distance for fixed class prototypes, or a scalar constant (here $d_{min}=1$) for learnable class prototypes.

\subsection{Generalization to Existing Object Detectors}
\label{sec:object_detector_integration}

In this section, we briefly revisit the two-stage, keypoint-based, and transformer-based object detectors that we use for the experiments with a special focus on their classification loss.

{\parskip=5pt
\noindent\textbf{Two-stage detectors}\label{sec:faster_rcnn_integration} such as Sparse R-CNN~\cite{sun2021sparse}, first extract a set of $N$ proposal boxes that potentially contain objects from latent image features. A RoIAlign operation extracts a fixed-sized feature vector for each proposal box. A \acrshort{roi} head then processes each proposal feature vector separately to extract classification features and bounding box regression values. In the case of Sparse R-CNN~\cite{sun2021sparse}, the \acrshort{roi} head takes the form of a dynamic instance interactive head, where each head is conditioned on a learned proposal feature vector. Sparse R-CNN~\cite{sun2021sparse} is trained end-to-end with a set prediction loss~\cite{Carion2020} that produces a cost-optimal bipartite matching between predictions and groundtruth objects. The classification accuracy enters the matching cost via a focal loss term. For our evaluations, we replace the classification head by the hyperbolic \acrshort{mlr} module as described in \secref{sec:hyperbolic}.}

{\parskip=5pt
\noindent\textbf{Keypoint-based detectors} \label{sec:centernet_integration} formulate object detection as a keypoint estimation problem. CenterNet~\cite{Zhou2019b} uses these keypoints as the center of bounding boxes. The output stage has a convolutional kernel with filter depth equal to the number of object classes and additional output heads for center point regression, bounding box width, and height, etc. Max-pooling on class-likelihood maps yields a unique anchor for each object under the assumption that no two bounding boxes share a center point. The class-likelihood maps are trained with a focal logistic regression loss, where each groundtruth bounding box is represented by a bivariate normal distribution $Y_{xyc}$ with mean at the groundtruth center point and class-specific variances.
We modify the classification heatmap to regress towards classification embeddings for each pixel. These embeddings are then transformed into hyperbolic space, and class heatmaps are generated by computing distance fields to each hyperbolic class prototype.}

{\parskip=5pt
\noindent\textbf{Transformer-based methods} were pioneered by the DETR~\cite{Carion2020} architecture that adopt an encoder-decoder topology to map a ResNet-encoded image features to a set of feature vectors by a transformer encoder-decoder network. Each feature vector is then independently decoded into prediction box coordinates and class labels by a 3-layer perceptron with ReLU activation~\cite{Carion2020}. Deformable DETR~\cite{Zhu2020} (DDETR) improves the computational efficiency of DETR by proposing a deformable attention module which attends only to a learned subset of sampling locations out of all the feature map pixels. The cross-attention modules in the neck decoder aggregate multi-scale feature maps, which facilitates the detection of small objects. The decoder consists of cross-attention modules that extract features as values, whereby the query elements are of N object queries and the key elements are of the output feature maps from the encoder. These are followed by self-attention modules among object queries.

We train the networks using a set prediction loss~\cite{Carion2020}, equivalent to that of Sparse R-CNN, described in \secref{sec:faster_rcnn_integration}.}

%% file: sections/experiments.tex
    \subsubsection{Datasets}
    
    
    
    \noindent\textit{COCO}~\cite{Lin2014} is currently the most widely used object detection dataset and benchmark. The images cover complex everyday scenes containing common objects in their natural context. Objects are labeled using per-instance segmentation to aid in precise object localization~\cite{Lin2014}.
    We use annotations for 80 “thing” objects types from the 2017 train/val split, with a total of 886,284 labeled instances in 122,266 images. The scenes range from dining table close-ups to complex traffic scenes.

    \noindent\textit{LVIS}~\cite{gupta2019lvis} builds on the COCO 2017 images but distinguishes 1203 object categories with a total of 127,0141 labeled instances in 100,170 training images alone. The class occurrences follow a discrete power law distribution, and is used as a benchmark for long-tailed object detection. 
    
    \noindent\textit{COCO 65/15} reuses the images and annotations from \acrshort{coco}, but holds out images with instances from 15 object types in the training set. We use the class selection as well as dataset-split from~\cite{Rahman2020}. 

    \subsubsection{Evaluation metrics} 
    We evaluate our proposed hyperbolic classification head for 2D object detection on the challenging \acrshort{coco} \textit{test-dev} and the long-tailed \acrshort{lvis} benchmark. Additionally, we evaluate the visual-to-semantic mapping performance on the zero-shot detection task using the classes split proposed in \cite{Rahman2020}. Our evaluation metric is the mean average precision ($mAP$) which defines the area under the precision-recall curve for detections averaged over thresholds for IoU $\in [0.5 : 0.05 : 0.95]$ (COCO’s standard metric). For closed-set object detection, we compare the mean over all classes, for long-tailed object detection on the \acrshort{lvis} dataset, and we also provide the mean of \underline{f}requent, \underline{c}ommon, and \underline{r}are classes. For zero-shot evaluation, we report average precision as well as recall for the 65 seen and the 15 unseen classes separately. We additionally report $AP_{cat}$, a modification of the average precision metric that defines a true positive detection if the \acrshort{iou} between the predicted and the groundtruth bounding box exceeds a threshold $\in [0.5 : 0.05 : 0.95]$ and is assigned the class label of any class with the same super-category in \acrshort{coco} stuff~\cite{caesar2018cvpr} label hierarchy as the groundtruth class. 
    

    \subsubsection{Training Protocol}
    

    
    We use the PyTorch~\cite{NEURIPS2019_9015} framework for implementing all the architectures, and we train our models on a system with an Intel Xenon@2.20GHz processor and NVIDIA TITAN RTX GPUs. All the methods use the ResNet-50 backbone with weights pre-trained on the ImageNet dataset~\cite{deng2009imagenet} for classification, and extract multi-scale feature maps using a \acrshort{fpn} with randomly initialized parameters. The hyperparameter settings, loss configuration, and training strategy principally follow the baseline configurations for maximum comparability, i.e., \cite{ren2015:faster-rcnn} for Faster R-CNN \acrshort{kge}, \cite{Zhou2019b} for CenterNet \acrshort{kge}, and \cite{Carion2020} for DETR \acrshort{kge} configurations. Please refer to the supplementary material for a detailed overview of hyperparameter settings and schedules. We train all the networks using an Adam optimizer with weight decay~\cite{loshchilov2017adamw} and gradient clipping. We use multi-scale training for all the models with a minimum training size of 480 pixel side length with random flipping and cropping for augmentation.

    \subsection{Benchmark Results}
    
    
    \subsubsection{\acrshort{coco} benchmark} 
    To compare the behaviors of hyperbolic classification heads with their baseline configurations, we evaluate the methods on the challenging \acrshort{coco} dataset. All object detectors were trained on 122,000 training images and tested on 5,000 validation images following the standard protocol~\cite{Zhou2019b,Zhu2020,sun2021sparse}.

    \input{tables/coco_val_results}
    \input{tables/coco_val_errors}
    
    \paragraph{Baselines:} We incorporate our hyperbolic classification heads into the two-stage, keypoint-based, and transformer-based object detector architectures a described in \secref{sec:object_detector_integration}. Consequently, we compare these methods against the standard configurations using a linear classification head in the euclidean embedding space.
    
    \paragraph{Discussions:}
    The results on the \acrshort{coco} \textit{val} set in \tabref{tab:coco_val_errors} indicate that hyperbolic
    classification heads consistently outperform their euclidean baseline configurations in the object detection performance. The Sparse R-CNN configuration achieves a substantial increase in the mean average precision of $+1.2\%$, without changes to the architecture and training strategy, only by modifying the algebra of the embedding space. The hyperbolic classification head's impact on various aspects of object detection are shown in \tabref{tab:coco_val_errors} for the \acrshort{coco} \textit{val} set. Surprisingly, the main gains from using the hyperbolic classification head are due to a consistent reduction in false positives, while the differences in other error metrics vary within $\pm 0.5\%$. Another noteworthy outcome is the consistent increase in $AP_{cat}$ when learning hyperbolic class prototypes.  

    Overall, these results indicate that hyperbolic classification heads perform superior to linear classification heads for the two-stage, keypoint-based, and transformer-based object detectors. The improvements are largest for the reduction of false positives and minor for localization errors. Interestingly, hyperbolic embeddings result in higher $AP_{cat}$, which suggest that it makes “better” detection errors as the misclassifications are still within the same supercategory and therefore more related to the true class.
    
    \subsubsection{\acrshort{lvis} benchmark} 
    The purpose of the \acrshort{lvis} experiments is to study the behavior of hyperbolic classification heads with a large set of object types and imbalanced class occurrences.  \tabref{tab:lvis_val_results} shows the results on the \acrshort{lvis} \textit{val} set for baseline methods as well as their counterparts using a hyperbolic classification head.

    \input{tables/lvis_val_results}

    \paragraph{Baselines:} 
    We compare our method against CenterNet2 using a federated loss~\cite{Zhou2021}, that computes a binary cross-entropy loss value over a subset of $|S|=50$ classes. The subset $S$ changes every iteration and is composed of all object types in the mini-batch's groundtruth and padded with randomly sampled negative classes. Additionally, we trained a Faster R-CNN model using a EQLv2 loss~\cite{tan2021equalization}, a mechanism to compensate class-imbalances in the dataset by equalizing the gradient ratio between positives and negatives for each class. 

    \paragraph{Discussions:}
         
    We observe a consistent improvement for the detection accuracy with fine-grained object types in hyperbolic embedding space, as both hyperbolic methods outperform their euclidean counterparts on precision for frequent classes $AP_f$. However, the hyperbolic classifiers perform inferior on rare and common classes. This effect is largest for the Faster R-CNN model trained by an EQLv2~\cite{tan2021equalization} loss. We therefore suggest that an improved class-balancing strategy needs to be designed for hyperbolic embeddings for long-tailed object detection, that needs to address both the class-imbalance by sampling and the impact of negatives have on the gradients.

    \subsubsection{Zero-shot evaluation} 
    Next, we assess the zero-shot abilities of hyperbolic embeddings on the \acrshort{coco} dataset using the \textit{65/15} classes split proposed by Rahman~\textit{et al.}~\cite{Rahman2020}.  Zero-shot object detection requires learning a mapping from visual to semantic feature space, such that the detector recognizes unseen object types only given their semantic representations. We investigate the behavior with semantic representations from word embeddings learned from the Wikipedia corpus by the \textit{word2vec} method~\cite{word2vec} in rely on the formulation for hyperbolic space by Leimeister~\textit{et al.}~\cite{leimeister2018skip}. Recent advances in zero-shot object detection rely on synthesizing unseen classes during training~\cite{hayat2020synthesizing}, or learn a reprojection of the semantic embedding vectors~\cite{Yan2022}. However, we reserve these tuning strategies of the network architecture and training pipeline to future work and focus on the straight-forward mapping from vision features to semantic class prototypes and a reprojection of embedding vectors as baselines. 
    
    \input{tables/coco_zeroshot_results}

    \paragraph{Baselines:} 
    We compare our method against object detectors using \textit{word2vec} word embeddings as class prototypes and trained with a polarity loss~\cite{Rahman2020} and a reconstruction loss~\cite{zheng2021zero}. The letter additionally learns a reprojection of semantic embedding vectors into the visual domain. We train our hyperbolic classifier using a focal loss~\cite{Lin2020}.
    
    \paragraph{Discussions:}
    The zero-shot performance of the hyperbolic classification head is shown in \tabref{tab:coco_zeroshot_unseen_recall}. The hyperbolic configurations outperform their naive baselines on average precision for seen and unseen classes, even though the baseline methods rely on more sophisticated training losses. The recall of groundtruth boxes appears to be dependent on the choice of loss function, as the Faster R-CNN baseline using the reconstruction loss proposed in \cite{zheng2021zero} achieves higher recall even though it yields the lowest precision on detecting unseen objects. The zero-shot performance using the hyperbolic Sparse R-CNN architecture shows superior results compared to all the baseline models and loss functions, which we take as an indication that the hyperbolic embedding space does not negatively affect the recall.

    \subsection{Hubness Analysis} 

\begin{figure}%
        \scriptsize
        \centering
        \begin{subfigure}[]{0.5\textwidth}
            \centering            \includegraphics[width=\textwidth]{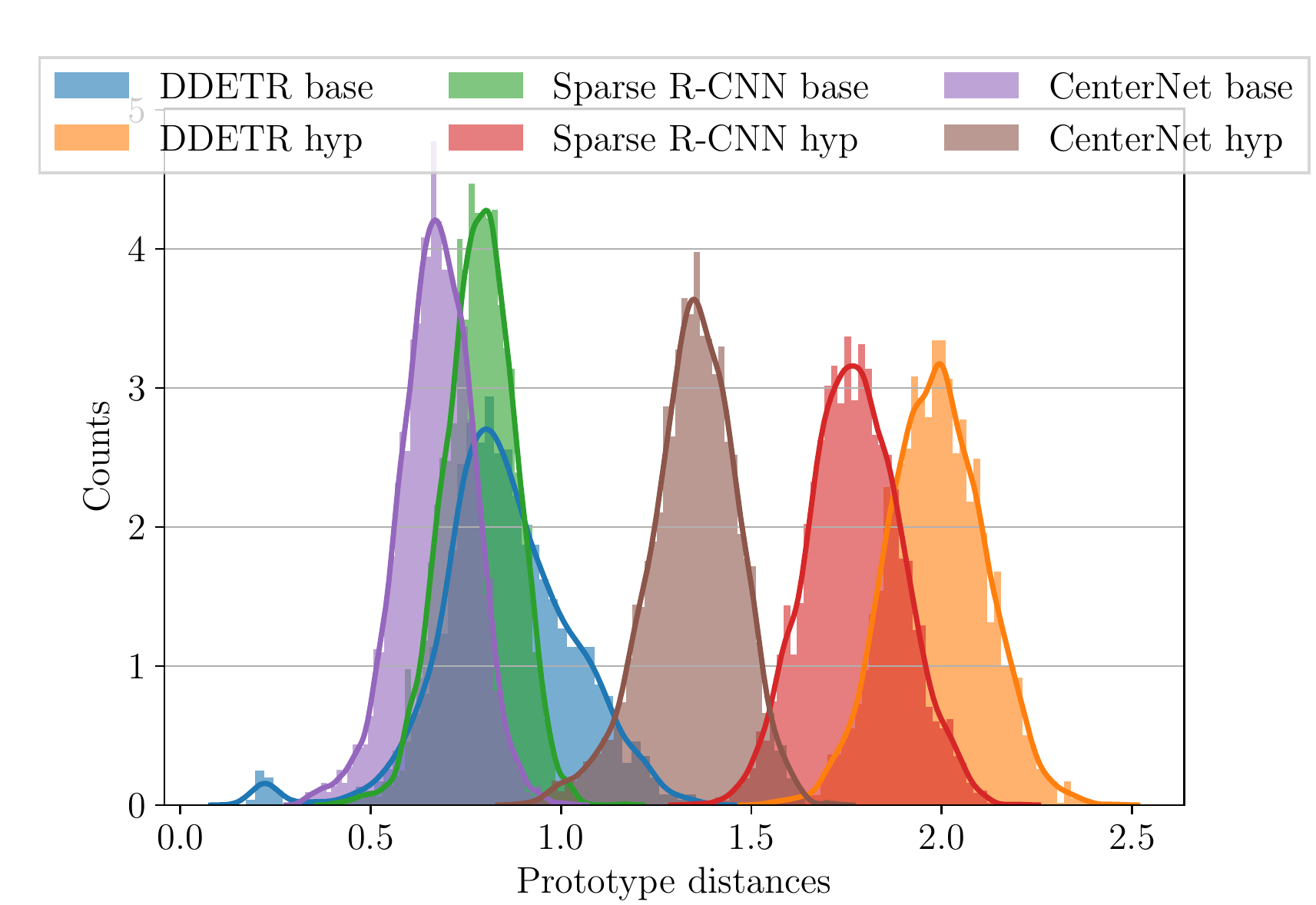}
        \end{subfigure}
        \hfill
        \begin{subtable}[]{0.45\textwidth}
            \centering
            \begin{tabular}[t]{llc}
                \toprule
                Model & Distance & $k_{skewness}$ \\
                \midrule
                CenterNet & Cossim & 2.50 \\
                CenterNet & Hyperbolic & 1.55 \\
                DDETR & Cossim & 2.51 \\
                DDETR & Hyperbolic & 1.85 \\
                Sparse R-CNN & Cossim & 2.18 \\
                Sparse R-CNN & Hyperbolic & 1.77 \\
                \bottomrule
            \end{tabular}
            \vspace{0.2cm}
        \end{subtable}%
        \caption{Comparison of \textit{k=5} nearest neighbors graph among class prototypes. (a) shows the histogram of all pairwise distances in the classification head, and (b) summarizes the skewness of the k-occurrence histograms as an indication of hubness.}%
        \label{fig:hubness_comparison}%
    \end{figure}

    The hubness problem is a phenomenon that can arise with nearest neighbor classification in high dimensional embedding spaces. It describes the emergence of hubs in the embedding space, i.e., a certain class prototype becoming the nearest neighbor of many samples, regardless of which class they belong to~\cite{fei2021z}. As a result, the histogram of the k-nearest neighbor graph is positively skewed. The literature proposes feature normalization strategies~\cite{fei2021z} to alleviate this effect. As we propose a novel distance metric in this work, we analyze its proneness to hubs in \figref{fig:hubness_comparison} and the need for such normalization strategies.
    
    The histograms on the left side in \figref{fig:hubness_comparison} show the pairwise distances between learned class prototypes in various detectors using a focal classification loss. For Deformable DETR (DDETR)~\cite{Zhu2020} and Sparse R-CNN~\cite{sun2021sparse}, we each considered the class prototypes of the output head. Focal loss assumes all classes to be distinct and therefore aims at maximizing inter-class distances. Hence, for linear classifiers in the euclidean space, it aims at pairwise orthogonal class prototypes, resulting in inter-class distances close to one. On the other hand, for hyperbolic distances, the distance measure is unbounded and thus the pairwise class distances vary noticeably.
    
    On the right side in \figref{fig:hubness_comparison}, we compare the skewness of the $k=5$ nearest neighbor graph. We observe that the $k=5$ nearest neighbor graphs of hyperbolic methods are consistently less positively skewed, even though the distances among class prototypes is larger on average. We therefore conclude that object detectors in hyperbolic space are less susceptible to the hubness problem and we can therefore skip the usage of a standardization technique.

    \subsection{Qualitative Results}

    In \figref{fig:coco_example}, we show qualitative detection results from two classifiers trained on the \acrshort{coco} \textit{train} set and classes. Detections on the left (a) were trained on object detection architectures using a learnable class prototypes in the euclidean space, while detections on the right (b) were generated by detectors using our proposed hyperbolic \acrshort{mlr} classifier. Interestingly, the accuracies of the two detectors are comparable, even though euclidean class prototypes resulted in a false positive in \figref{fig:coco_example}~(a). Another noticeable difference is the composition of the top-3 class scores that provide an insight in the structure of the embedding spaces. While most bounding boxes were classified correctly by both classifiers, the learned hyperbolic embeddings appear to have grouped categorically similar concepts as the detector predicts classes of \textit{vehicle} types for the \textit{car} object instance. For the \textit{person} instance, there are no equivalent categorical classes, so there seems to have emerged a \textit{living thing} neighborhood and a neighborhood of frequently co-occurring classes containing \textit{bicycle} in the embedding space.
    
    \input{graphics/display_coco_results}
    \input{graphics/display_lvis_results}
    \input{graphics/display_coco_zeroshot_results}
        
    The example predictions for \acrshort{lvis} classes are shown in \figref{fig:lvis_example}. The hyperbolic method recognizes even partly visible objects and top-3 predictions are more semantically similar, such as \textit{headband} and \textit{bandanna} for the child in the center. This indicates that the embeddings capture more conceptual similarities than the euclidean classifier. However, “rare” classes are missing from the top-3 predictions, as a result the hyperbolic method fails to assign the correct class \textit{ballet skirt} for the tutu worn by the child on the right.
    
    \figref{fig:coco_zeroshot_example} presents the zeros-hot results for an unseen \textit{airplane} instance. Here, the hyperbolic model also yields categorically similar predictions when given semantic word embeddings. Surprisingly, this is not as strong for the euclidean classifier (a), even though it maps the visual features to word embeddings. However, we note that the confidence by the hyperbolic method (i.e. distance in embedding space) is considerably lower (larger) for the unseen class and the \textit{airplane} bounding box appears less accurate. This could be mitigated by using more sophisticated zero-shot pipelines such as the reconstruction loss~\cite{zheng2021zero} or synthetic training samples~\cite{hayat2020synthesizing}.

%% file: tables/coco_val_results.tex
\begin{table}[t]
\centering
\footnotesize
\caption{COCO 2017 \textit{val} results for methods using linear layers (top row) compared to hyperbolic (bottom row) classification heads.}
\begin{tabular}{p{4.8cm}p{0.8cm}p{0.8cm}p{0.8cm}p{0.8cm}p{0.8cm}p{0.8cm}}
\toprule
Method & ${\text{AP}}$ & ${\text{AP}_{50}}$ & ${\text{AP}_{75}}$ & ${\text{AP}_{s}}$ & ${\text{AP}_{m}}$ & ${\text{AP}_{l}}$ \\ \hline
CenterNet (R-50-FPN)~\cite{Zhou2019b} & 40.2 & 58.2 & 43.7 & 23.2 & 44.6 & 52.3 \\
+ Hyperbolic classifier & 41.1 & 59.3 & 45.2 	& 24.2 	&45.0 	&53.8 \\ \midrule
DDETR (R-50-FPN)~\cite{Zhu2020} & 44.5 & 63.5 & 48.7 & 26.8 & 47.7 & 59.5 \\
 + Hyperbolic classifier & 45.0 	&64.3 	&48.8 	&27.9 	&47.9 	&59.7 \\ \midrule
Sparse R-CNN (R-50-FPN)~\cite{sun2021sparse} & 45.0 & 64.1 & 49.0 & 27.8 & 47.5 & 59.7 \\
 + Hyperbolic classifier & \textbf{46.2} & \textbf{65.9} & \textbf{50.7} & \textbf{29.2} & \textbf{49.2} & \textbf{60.7} \\ \bottomrule
\end{tabular}
\label{tab:coco_test_results}
\vspace{-0.4cm}
\end{table}

%% file: tables/coco_val_errors.tex
\begin{table}[t]
    \centering
    \footnotesize
    \caption{COCO 2017 \textit{val} results using single-scale testing. Each section compares the results given identical networks but a linear classification head (top row) with a Hyperbolic classification head (bottom row). The error metrics $E_x$ were computed as proposed by Boyla \textit{et al.} \cite{bolya2020tide}.}
    \begin{tabular}{lp{1.0cm}p{1.2cm}p{0.8cm}p{0.8cm}p{0.8cm}p{0.8cm}p{0.8cm}p{0.8cm}}
    \toprule
    Method                            & $mAP$ & $mAP_{cat}$ & $E_{cls}$ & $E_{loc}$ & $E_{bkg}$ & $E_{miss}$ & $E_{FP}$ & $E_{FN}$ \\ \midrule
    CenterNet (R-50-FPN)~\cite{Zhou2019b}                  & 40.2  & 38.8        & 3.1       & \textbf{6.1}       & 3.7       & 5.4        & 21.7     & 12.6     \\
     + Hyperbolic classifier &   41.1 & 39.5 & 2.9  & 5.7  & 3.7  & 5.4  & 21.4 & 11.9 \\ \midrule
    DDETR (R-50-FPN)~\cite{Zhu2020} &   44.5    & 40.7    &  2.3   & 7.5  &  4.2 & 4.4 & 19.6  & 12.2 \\
     + Hyperbolic classifier & 45.0  & \textbf{47.1}  & 2.4  & 7.5  & 4.2  & \textbf{4.0} & 19.0	& 11.4 \\ \midrule
    Sparse R-CNN (R-50-FPN)~\cite{sun2021sparse} & 45.0  & 43.0        & \textbf{2.1}       & 6.7       & 3.9       & 4.7        & 19.6     & \textbf{11.2}     \\
     + Hyperbolic classifier          &   \textbf{46.2}  & 45.6  & 2.4 & 6.8 & \textbf{3.6} & 5.0 & \textbf{17.6} & 11.7  \\ \bottomrule
    \end{tabular}
    \label{tab:coco_val_errors}
\end{table}

%% file: tables/lvis_val_results.tex
\begin{table}[t]
\centering
\footnotesize
\caption{LIVS \textit{val} results for various models with linear classification head and hyperbolic classification head.
All methods were trained by repeat factor resampling~\cite{gupta2019lvis} by a factor of $0.001$. 
}
\begin{tabular}{lp{1.6cm}p{0.8cm}p{0.8cm}p{0.8cm}p{0.8cm}p{0.8cm}p{0.8cm}}
\toprule
Model                    & Loss      & $AP$     & $AP_{50}$ & $AP_{75}$ & $AP_{r}$ & $AP_{c}$ & $AP_{f}$ \\ \midrule
CenterNet2   (R-50-FPN)~\cite{Zhou2021}  &  FedLoss  &  \textbf{28.2}    &    39.2   &	\textbf{30.0}    & \textbf{18.8}     & \textbf{26.4}     & 34.4     \\
 + Hyperbolic classifier &  FedLoss  &  27.9    &  39.7      &  28.8     & 17.3    & 26.2   & \textbf{34.9}   \\ \midrule
Faster R-CNN (R-50-FPN)~\cite{wu2019detectron2} &  EQLv2    &   23.6   &  39.3     &  24.5     &  14.2    &  22.3    &  29.1        \\
 + Hyperbolic classifier &  EQLv2    &   23.2   & \textbf{40.1}    & 24.7      & 11.9     & 21.0    &    29.5 \\ \bottomrule
\end{tabular}
\label{tab:lvis_val_results}
\vspace{-0.4cm}
\end{table}

%% file: tables/coco_zeroshot_results.tex
\begin{table}[t]
\centering
\footnotesize
\caption{Precision and recall on seen as well as unseen classes for COCO 2017 65/15 split. An \acrshort{iou} threshold of 0.5 is used for computing recall and average precision. \textit{HM} refers to the harmonic mean of seen and unseen classes.}
\begin{tabular}{lllcccccc}
\toprule
\multirow{2}{*}{Model} & \multirow{2}{*}{Method} & \multirow{2}{*}{Embedding} & \multicolumn{2}{c}{Seen} & \multicolumn{2}{c}{Unseen} & \multicolumn{2}{c}{HM} \\
\cmidrule(lr){4-5}\cmidrule(lr){6-7}\cmidrule(lr){8-9}
\multicolumn{1}{c}{} & & & mAP & Recall & mAP & Recall & mAP & Recall \\ \midrule
RetinaNet & PL-vocab \cite{rahman2020improved} & word2vec & 34.1 & 36.4 & 12.4 & 37.2 & 18.2 & 36.8 \\
Faster R-CNN & ZSI \cite{zheng2021zero} & word2vec & 35.8 & 62.6 & 10.5 & 50.0 & 16.2 & 55.6 \\
Faster R-CNN & Hyperbolic & word2vec & 36.2 & 52.4 & 13.1 & 40.5 & 19.3 & 45.7 \\
Sparse R-CNN & Hyperbolic & word2vec & \textbf{38.0} & \textbf{65.4} & \textbf{13.5} & \textbf{50.4} & \textbf{19.9} & \textbf{56.9} \\ \bottomrule
\end{tabular}
\label{tab:coco_zeroshot_unseen_recall}
	\vspace{-0.4cm}
\end{table}

%% file: graphics/display_coco_results.tex
\begin{figure}[t]
    \centering
    \subfloat[\centering Learned euclidean embeddings.]{{\includegraphics[width=0.49\textwidth]{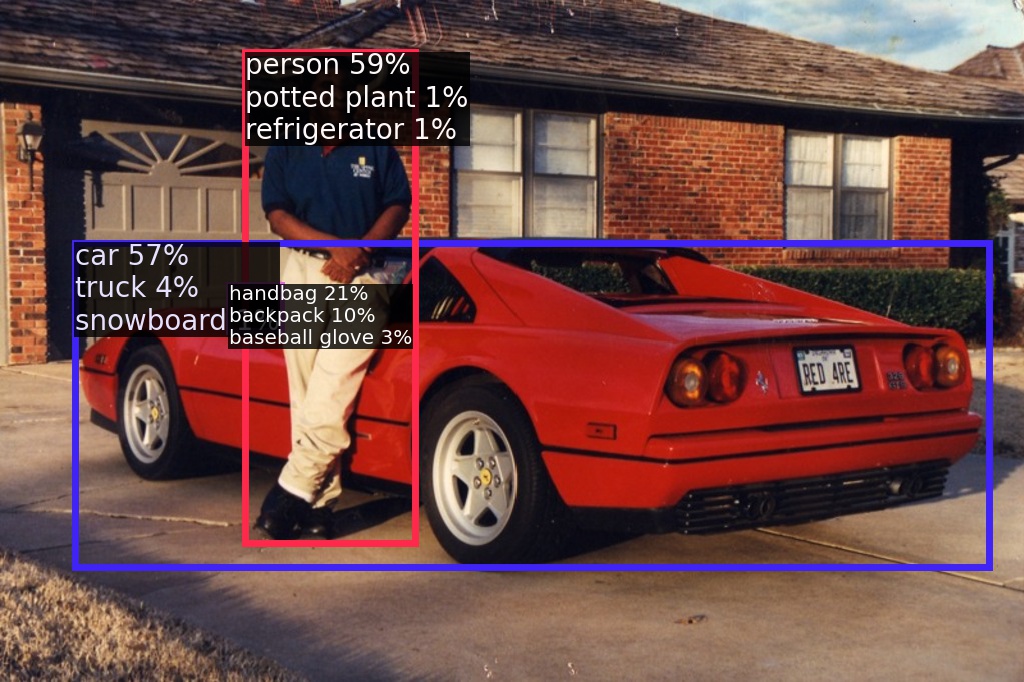} }}%
    \hfill
    \subfloat[\centering Learned hyperbolic embeddings.]{{\includegraphics[width=0.49\textwidth]{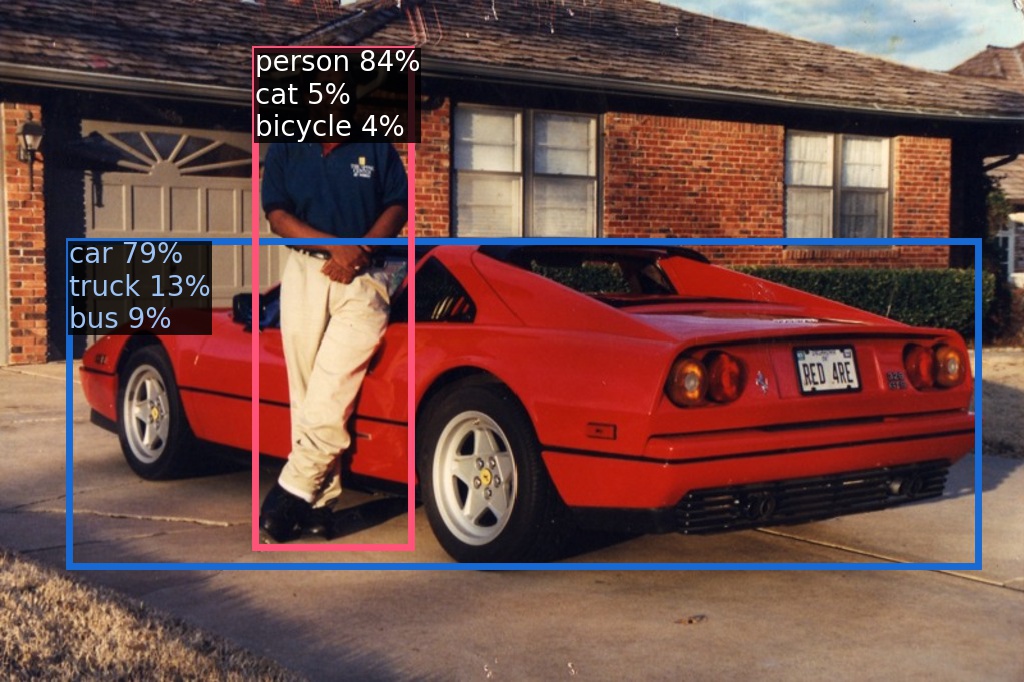} }}%
    \caption{Qualitative results for Sparse R-CNN model trained on full \acrshort{coco} classes with an (a) euclidean and (b) hyperbolic classification head.}%
    \label{fig:coco_example}%
	\vspace{-0.4cm}
\end{figure}

%% file: graphics/display_lvis_results.tex
\begin{figure}[t]
    \centering
    \subfloat[\centering Learned euclidean embeddings.]{{\includegraphics[trim=0 20 0 20,clip,width=0.49\textwidth]{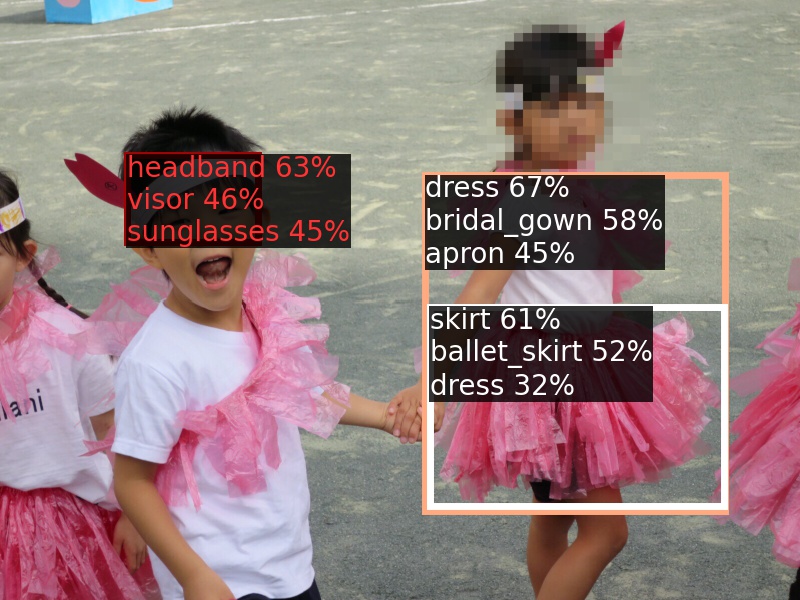} }}%
    \hfill
    \subfloat[\centering Learned hyperbolic embeddings.]{{\includegraphics[trim=0 20 0 20,clip,width=0.49\textwidth]{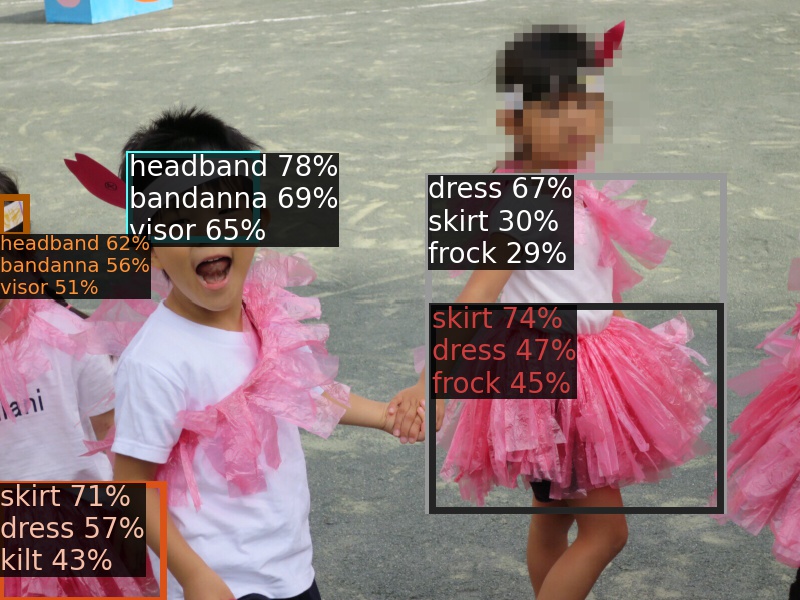} }}%
    \caption{Qualitative results for CenterNet2 model trained on \acrshort{lvis} classes  with an (a) euclidean and (b) hyperbolic classification head. The children are wearing \textit{ballet skirt}s, a rare class in the \acrshort{lvis} dataset with $<10$ training samples.}%
    \label{fig:lvis_example}
	\vspace{-0.2cm}
\end{figure}


%% file: graphics/display_coco_zeroshot_results.tex
\begin{figure}[t]
    \centering
    \subfloat[\centering word2vec~\cite{word2vec} euclidean embeddings.]{{\includegraphics[trim=80 70 100 70,clip, width=0.49\textwidth]{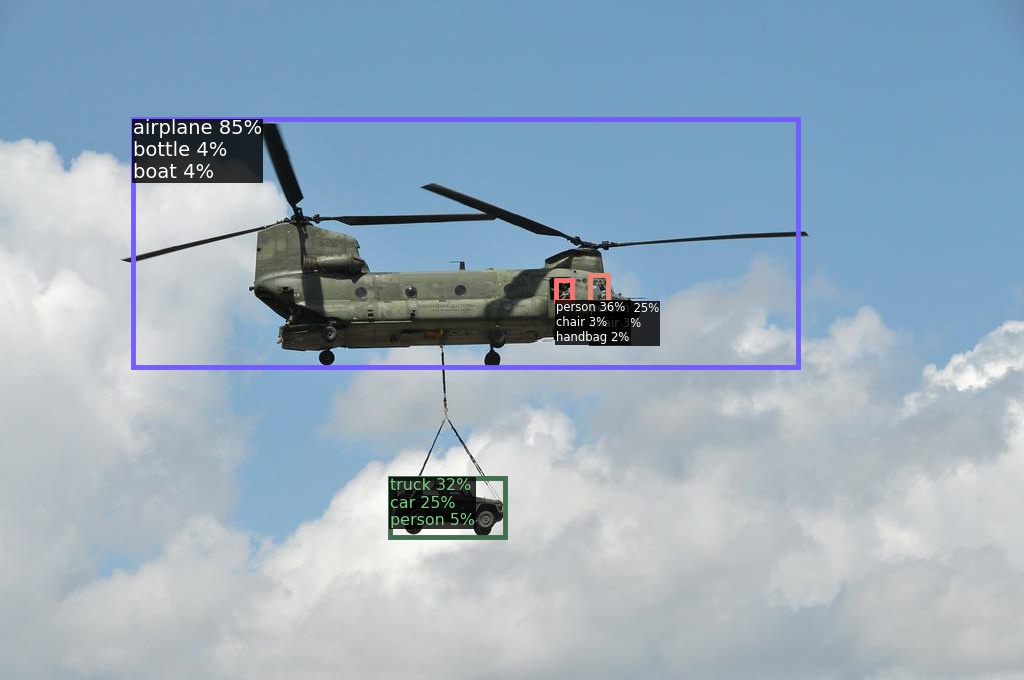} }}%
    \hfill
    \subfloat[\centering word2vec~\cite{leimeister2018skip} hyperbolic embeddings.]{{\includegraphics[trim=80 70 100 70,clip, width=0.49\textwidth]{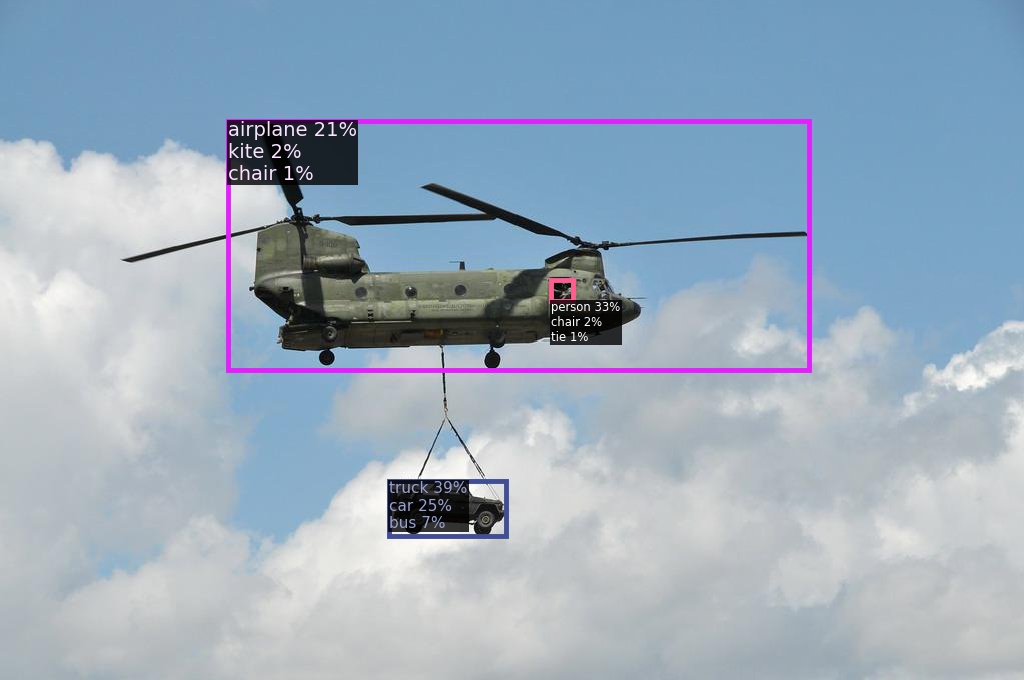} }}%
    \caption{Qualitative results for a Faster R-CNN model trained on the seen classes of the \acrshort{coco} 65-15 split with an (a) euclidean and (b) hyperbolic classification head on word2vec semantic embeddings. The image shows a seen \textit{car} instance as well as an unseen \textit{airplane} instance.}%
    \label{fig:coco_zeroshot_example}%
	\vspace{-0.4cm}
\end{figure}


%% file: sections/conclusion.tex
In this work, we proposed to use hyperbolic embedding spaces to learn class prototypes in object detection.
We extended two-stage, keypoint-based, and transformer-based object detectors to incorporate hyperbolic classification heads.
Evaluations on closed-set, long-tailed, as well as zero-shot settings showed that the hyperbolic methods outperformed their euclidean baselines on classes with sufficient training samples.
The hyperbolic classification heads resulted in considerably fewer false positives, and produced “better” classification errors, misclassified labels were within the same supercategory as the true classes.
We therefore conclude that hyperbolic geometry provides a promising embedding space for object detection and deserves future work to design optimized class-balancing and zero-shot training frameworks.

%% file: sections/supplementary.tex
\title{On Hyperbolic Embeddings in Object Detection  \\ \textit{Supplementary Material}}

\titlerunning{On Hyperbolic Embeddings in 2D Object Detection}
%
\author{Christopher Lang\inst{1,2} \and
Alexander Braun\inst{1} \and
Abhinav Valada\inst{2}}
\authorrunning{C. Lang et al.}
%
\institute{Robert Bosch GmbH
\and
University of Freiburg\\
\email{\{lang,valada\}@cs.uni-freiburg.de}}
\maketitle

\appendix

\section{Detailed training settings}

The training settings for are shared for all configurations of the same base architectures. All experiments were performed using a AdamW~\cite{loshchilov2017adamw} optimizer with $10^{-4}$ weight decay as well as gradient clipping at a norm of $0.1$. Detailed training configurations per base architecture for experiments on the \acrshort{coco} dataset are given in \tabref{tab:training_settings_coco}, and for experiments on \acrshort{lvis} in \tabref{tab:training_settings_lvis} respectively.

\begin{table}[h]
\centering
\footnotesize
\caption{Detailed training settings for networks trained on \acrshort{coco} \textit{train} dataset. }
\label{tab:training_settings_coco}
    \begin{tabular}{p{2.8cm}|p{3.5cm}p{3.5cm}p{2.5cm}} \toprule
    Model & Sparse R-CNN~\cite{sun2021sparse} & CenterNet~\cite{Zhou2019b} & DDETR~\cite{Sun2020} \\ \midrule
    Backbone & R50-FPN & R50-FPN & R50-FPN \\
    Prototype dim & 1024 & 64 & 256 \\
    Min train size & 640 & 640 & 480 \\
    Max train size & 1333 & 640 & 1033 \\
    Augmentations & horizontal flip & horizontal flip & \thead[l]{\footnotesize horizontal flip,\\+ random resize\\+  random crop} \\
    Batch size & 16 & 16 & 16 \\
    Optimizer & AdamW & AdamW & AdamW \\
    Train epochs & 40 & 58 &  \\
    Base LR & 0.02 & 0.01 & $10^{-5}$ \\
    Scheduler & \thead[l]{\footnotesize Linear warmup\\+ Multi-step schedule} & \thead[l]{\footnotesize Linear warmup\\+ Multi-step schedule} &  Linear warmup\\
    Warmup iters & 1000 & 4000 & 10 \\ 
    \thead[l]{\footnotesize LR reduction\\ {[epochs]}} & (25,35) & (38,52) & - \\
    Gradient clip norm & 1.0 & 1.0 & 0.1 \\\bottomrule
    \end{tabular}
\end{table}

\begin{table}
\footnotesize
\centering
\caption{Detailed training settings for networks trained on \acrshort{lvis} \textit{train} dataset. Methods were trained with repeated factor sampling~\cite{gupta2019lvis} by a repeat factor of $10^{-3}$}
\label{tab:training_settings_lvis}
    \begin{tabular}{p{3cm}|p{4cm}p{4cm}} \toprule
    Model & Faster R-CNN & CenterNet2 \\ \midrule
    Prototype dim & 1024 & 1024 \\
    Min train size & 640 & 640 \\
    Max train size & 1333 & 1333 \\
    Augmentations & horizontal flip & horizontal flip \\
    Batch size & 16 & 16 \\
    Optimizer & AdamW & AdamW \\
    Train epochs & 29 & 15 \\
    Base LR & 0.02 & 0.02 \\
    LR reductions steps & (19,24) & (10,13) \\
    Scheduler & \thead[l]{Linear warmup\\+ Multi-step schedule} & \thead[l]{Linear warmup\\+ Multi-step schedule}  \\
    Warmup iters & 1000 & 4000 \\ \bottomrule
    \end{tabular}
\end{table}

\section{\acrshort{coco} \textit{test-dev} results}

In \tabref{tab:coco_test_results}, we summarize the results on the \acrshort{coco} \textit{test-dev} benchmark for CenterNet~\cite{Zhou2019b}, Deformable DETR~\cite{Zhu2020}, and Sparse R-CNN~\cite{sun2021sparse} in their baseline configuration using a euclidean classification head, as well as using a hyperbolic classification head based on the hyperboloid model.

\input{tables/coco_test_results}

\paragraph{Discussion:}
The results on the \acrshort{coco} \textit{test-dev} set are shown in \tabref{tab:coco_test_results}. The hyperbolic classification heads perform superior to their baselines for both the CenterNet and the Sparse R-CNN method.  The Sparse R-CNN configuration achieves a substantial increase in the mean average precision of $+1.1\%$ in comparison to its euclidean baseline, and especially boosts performance on small objects by $+3.0\% AP_s$. 
Even though the DDETR method using a hyperbolic head outperforms its linear classifier baseline on the \acrshort{coco}~\textit{val} set, it cannot match the baseline's unusual $+2.4\% AP$ jump from ~\acrshort{coco}~\textit{val} to \acrshort{coco}~\textit{test-dev}. 
As reported in Tab. 2 of the main paper, the false positives are the strongest impairment for all object detectors to a perfect $AP$ performance. Therefore, measures to reduce false positives in the training hyperparameters appear promising to further improve performance of hyperbolic classifiers on \acrshort{coco} \textit{test-dev}.

%% file: tables/coco_test_results.tex
\begin{table}
\centering
\footnotesize
\caption{COCO 2017 test-dev results for methods using euclidean classification heads (top row) compared to hyperbolic classification heads (bottom row in each section).}
\label{tab:coco_test_results}
\begin{tabular}{p{4.8cm}p{0.8cm}p{0.8cm}p{0.8cm}p{0.8cm}p{0.8cm}p{0.8cm}}
\toprule
Method & ${\text{AP}}$ & ${\text{AP}_{50}}$ & ${\text{AP}_{75}}$ & ${\text{AP}_{s}}$ & ${\text{AP}_{m}}$ & ${\text{AP}_{l}}$ \\ \hline
CenterNet (R-50-FPN)~\cite{Zhou2019b} & 40.4 & 58.6 & 44.1 & 22.8 & 43.7 & 50.8 \\
+ Hyperbolic classifier & 40.7 & 58.8 & 44.2 & 25.9 & 42.6 & 50.3 \\ \midrule
DDETR (R-50-FPN)~\cite{Zhu2020} & 46.9 & 66.4 & 50.8 & 27.7 & 49.7 & 59.9 \\
 + Hyperbolic classifier & 45.3 & 64.1 & 49.3 & 26.1 & 48.1 & 57.8 \\ \midrule
Sparse R-CNN (R-50-FPN)~\cite{sun2021sparse} & 45.2 & 64.6 & 49.3 & 26.9 & 47.4 & 57.4 \\
 + Hyperbolic classifier & 46.3 & 65.5 & 51.9 & 29.9 & 48.5 & 57.2 \\ \bottomrule
\end{tabular}
\end{table}